%% file: acl.tex
\pdfoutput=1

\documentclass[11pt]{article}

\usepackage{acl}

\usepackage{times}
\usepackage{latexsym}

\usepackage[T1]{fontenc}

\usepackage[utf8]{inputenc}

\usepackage[ruled]{algorithm2e}
\usepackage{microtype}
\usepackage{booktabs}
\usepackage{hyperref}
\usepackage{url}
\usepackage{amsmath}
\usepackage{graphicx}
\usepackage{subcaption}

\definecolor{mygray}{gray}{.9}

\title{Memformer: A Memory-Augmented Transformer for Sequence Modeling}

\author{
    Qingyang Wu \textsuperscript{\rm 1},
    Zhenzhong Lan \textsuperscript{\rm 2},
    Kun Qian \textsuperscript{\rm 1}
    Jing Gu \textsuperscript{\rm 1} \\
    \textbf{Alborz Geramifard} \textsuperscript{\rm 3}
    \textbf{Zhou Yu} \textsuperscript{\rm 1} \\
    \textsuperscript{\rm 1} University of California, Davis,
    \textsuperscript{\rm 2} Westlake University, \textsuperscript{\rm 3} Facebook AI \\
    \texttt{\{wilwu,qkun,jkgu,joyu\}@ucdavis.edu}  \\ \texttt{lanzhenzhong@westlake.edu.cn,alborzg@fb.com} \\
}
\begin{document}
\maketitle
\begin{abstract}

Transformers have reached remarkable success in sequence modeling. 
However, these models have efficiency issues as they need to store all the history token-level representations as memory.
We present Memformer, an efficient neural network for sequence modeling, that utilizes an external dynamic memory to encode and retrieve past information.
Our model achieves linear time complexity and constant memory space complexity when processing long sequences.
We also propose a new optimization scheme, memory replay back-propagation (MRBP), which promotes long-range back-propagation through time with a significantly reduced memory requirement.
Experimental results show that Memformer has achieved comparable performance compared against the baselines by using 8.1x less memory space and 3.2x faster on inference.
Analysis of the attention pattern shows that our external memory slots can encode and retain important information through timesteps.

\end{abstract}

\input{introduction.tex}

\input{related_work.tex}

\input{methods.tex}

\input{experiments.tex}
\input{conclusion.tex}

\newpage

\bibliography{anthology,custom}
\bibliographystyle{acl_natbib}

\onecolumn
\appendix

\section{MRBP Efficiency Test}

In this section, we test MRBP's efficiency by comparing against the standard back-propagation through time (BPTT) and the standard gradient checkpointing (GC) algorithm.
This algorithm is useful for Memformer to reduce memory requirement because of the back-propagation through several timesteps.
We use the Memformer model and set all the hyper-parameters to be the same.

\begin{table}[h]
    \centering
    \resizebox{0.470\textwidth}{!}{
        \begin{tabular}{l|cc}
            \toprule
            Method & GPU Memory (MB) & Speed (relative) \\
            \midrule
            BPTT   & 16,177          & x1.00            \\
            GC     & 9,885           & x0.48            \\
            MRBP   & 7,229           & x0.90            \\
            \bottomrule
        \end{tabular}
    }
    \caption{Memory Replay Back-Propagation performance comparison. Evaluation speed is based on seconds per sample. BPTT means back-propagation through time. GC means gradient checkpointing.}
\end{table}

The back-propagation through time (BPTT) approach is the fastest because it does not need re-computation.
However, it costs the most amount of memory due to unrolling the entire computational graph.
While gradient checkpointing can save huge amount of memory, it is much slower than the other two methods (x0.48).
In contrast, our MRBP saves more GPU memory with only slight speed degeneration (x0.90).

\section{Training Details}

\begin{table}[h]
    \centering
    \resizebox{0.6\textwidth}{!}{
        \begin{tabular}{l|cc}
            \toprule
                              & Image Generation & Language Modeling \\
            \midrule
            batch size        & 256              & 128               \\
            warm-up steps     & 1,000            & 1,0000            \\
            learning rate      & 1e-3             & 1e-3              \\
            dropout           & 0.1              & 0.1               \\
            memory length     & 8                & 1,024             \\
            temperature       & 0.25             & 0.125             \\
            time horizon      & 8                & 8                 \\
            weight decay      & 0.01             & 0.01              \\
            max gradient norm & 1.0              & 1.0               \\
            training steps    & 10,000           & 150,000           \\
            \bottomrule
        \end{tabular}
    }
    \caption{Training Details}
\end{table}

We trained our model on NVIDIA V100 16GB and 2080Ti 11GB.
The training for image generation took about one day on one GPU.
The training for language modeling took approximately four days on four GPUs.

\section{Effects of Time Horizon and Memory Size}

\begin{figure}[h]
    \centering
    \begin{subfigure}{0.4\textwidth}
        \centering
        \includegraphics[width=\linewidth]{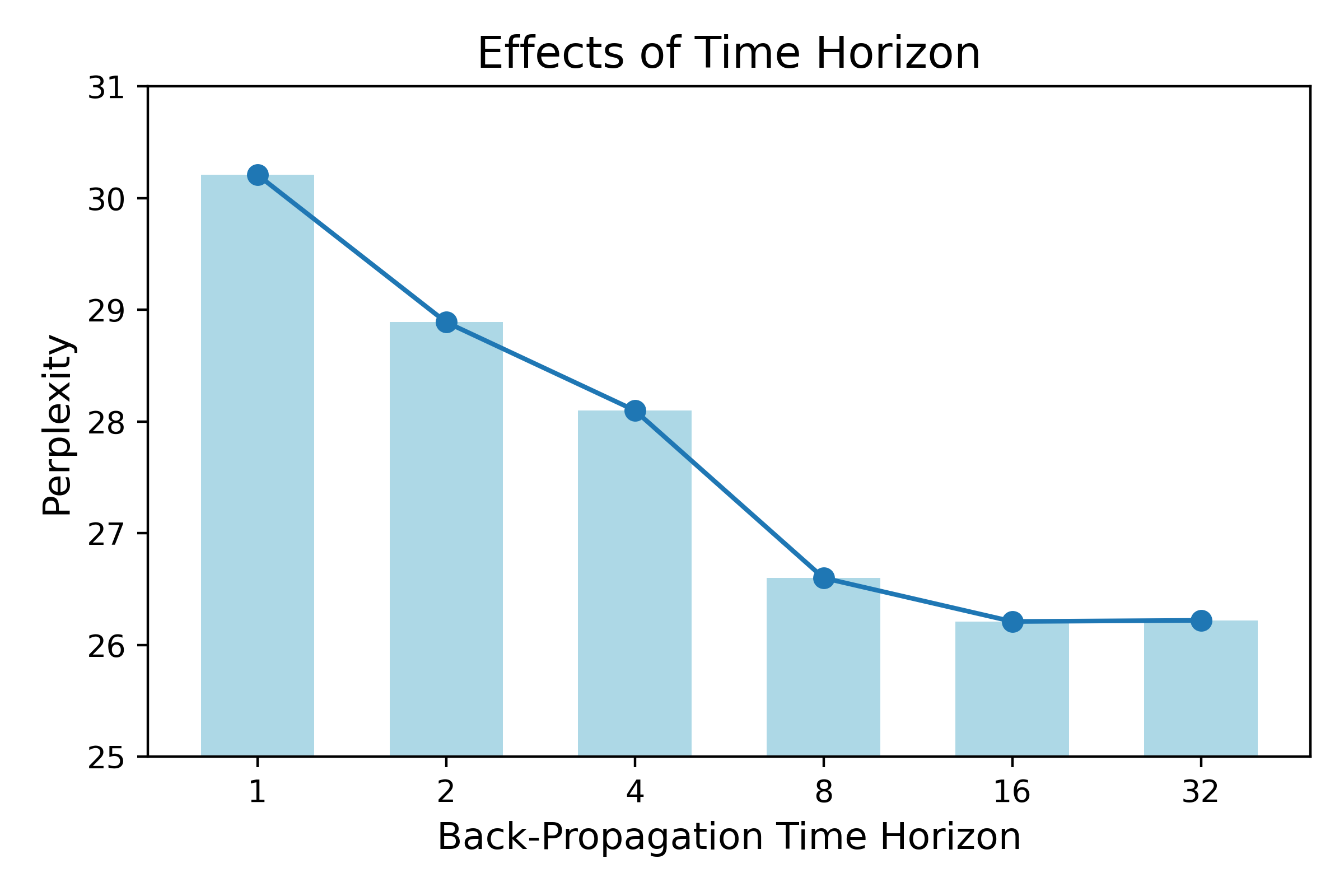}
        \caption{Effects of different time horizons}
        \label{fig:time_horizon}
    \end{subfigure}
    \begin{subfigure}{0.4\textwidth}
        \centering
        \includegraphics[width=\linewidth]{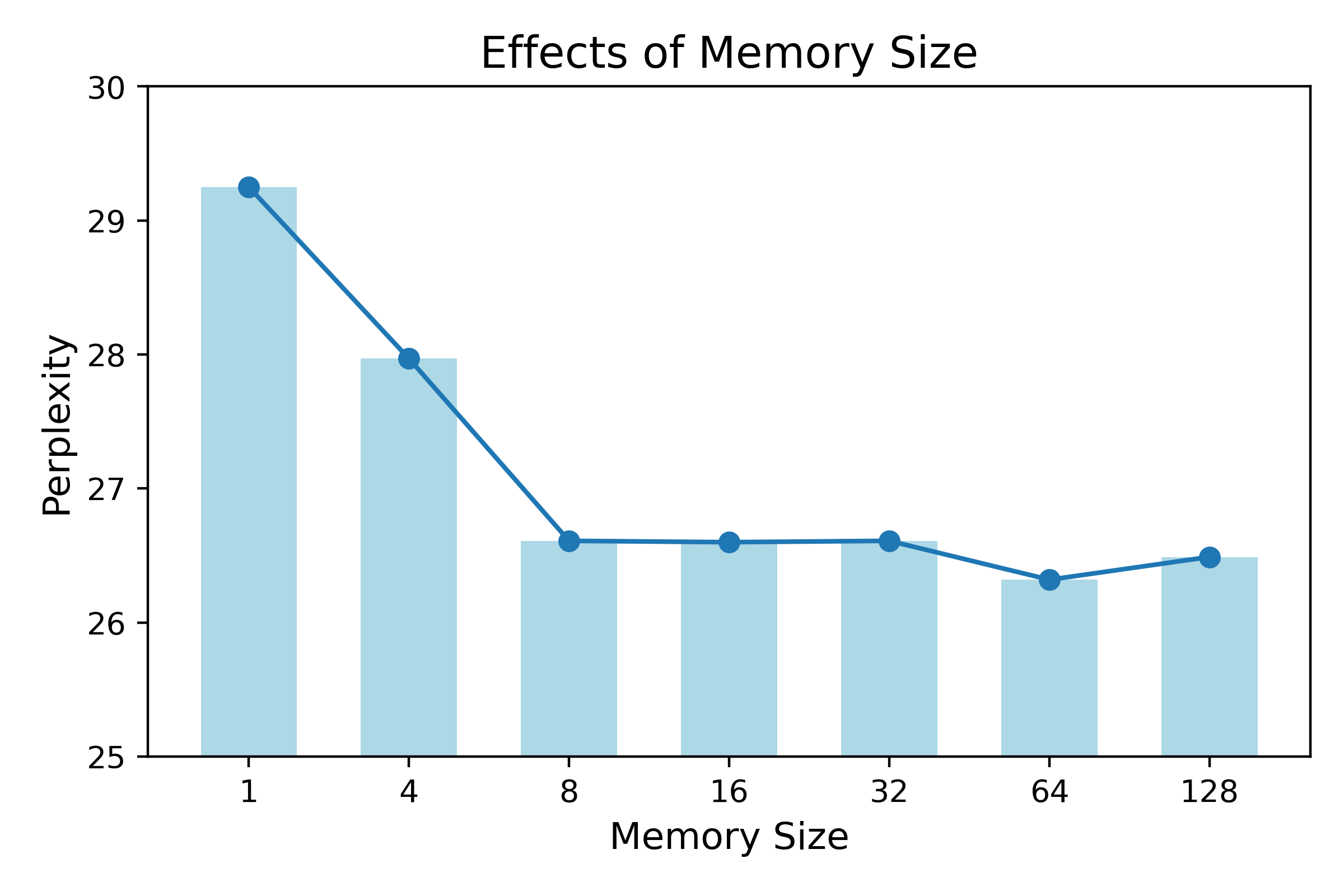}
        \caption{Effects of different memory sizes}
        \label{fig:memory_size}
    \end{subfigure}
    \caption{Effects of different configurations. (a) shows the effects of changing time horizon. (b) shows the effects of changing memory size.}
\end{figure}

We test how the time horizon for back-propagation affects the performance.
We test on a smaller Memformer model for the efficiency.
The results are shown in Figure~\ref{fig:time_horizon}.
We vary the back-propagation time horizon from 1 to 32.
When the time horizon is set to 1, back-propagation cannot pass gradients through memory to the previous timestep.
Thus, we observe the performance is the worst when the time horizon is 1.
As we increase the time horizon, the model achieves better perplexity scores.
When the time horizon is increased to 32, we observe the marginal improvement on perplexity is almost gone.
A large memory size ideally helps to store more information.
From Table~\ref{fig:memory_size}, we can see a huge improvement when increasing the memory size from 1 to 8.
Furhter increasing the memory size has a smaller effects on the performance, and we suspect that this is due to the size of the model.

\section{Implementation of Memory Writer}

\begin{figure}[h]
    \centering
    \includegraphics[width=0.6\textwidth]{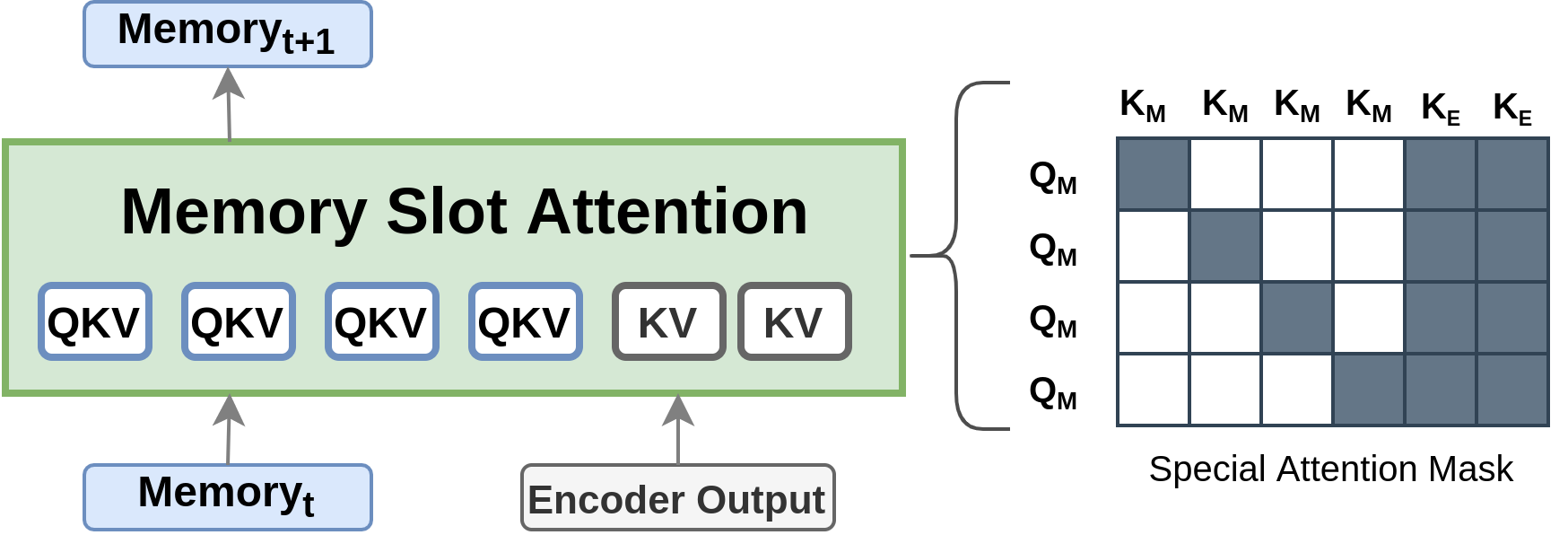}
    \caption{Memory Writer's Attention}
    \label{fig:mem_slot_attn}
\end{figure}

Memory Slot Attention in Figure~\ref{fig:mem_slot_attn} produces the next timestep's memory $M_{t+1}$.
This module takes the inputs of the previous timestep's memory $M_{t}$ and the encoder's final hidden states.
It then projects the memory into queries, keys, and values, while the encoder outputs are into keys and values.
Since each memory slot should not be interfering with other memory slots, we design a special type of sparse attention pattern.
Thus, each slot in the memory can only attend over itself and the encoder outputs.
This is to preserve the information in each slot longer over the time horizon.
For example, if one slot only attends  itself, then the information in that slot will not change in the next timestep.

\end{document}

%% file: introduction.tex
\section{Introduction}

Memory plays a fundamental role in human cognition.
Humans perceive and encode sensory information into a compressed representation stored in neurons, 
and later we effectively retrieve the stored information to accomplish various tasks.
The formation of memory involves complex cognitive processes.
Modeling and studying the behavior of human memory is still a challenging research problem in many areas.

Many researchers have attempted to incorporate memory systems in artificial neural networks.
Early works like recurrent neural networks (RNN) \citep{10.5555/65669.104451} including LSTM \citep{DBLP:journals/neco/HochreiterS97} and GRU \citep{69e088c8129341ac89810907fe6b1bfe} model temporal sequences with their internal compressed state vector as memory.
However, they are limited in preserving the long-term information due to the memory bottleneck.
To alleviate this limitation, more powerful memory network architectures such as Neural Turing Machine (NTM) \citep{DBLP:journals/corr/GravesWD14}, Differential Neural Computer (DNC) \citep{DBLP:journals/nature/GravesWRHDGCGRA16} have been proposed by leveraging a large external dynamic memory. 
Unfortunately, due to their complex memory interaction mechanism, they are not widely used for down-stream tasks at present.

More recently, \citet{DBLP:conf/nips/VaswaniSPUJGKP17} propose Transformer by discarding the use of recurrence and memory.
Instead, it computes all the $\mathcal{O}(N^2)$ paired dependencies in a sequence with self-attention \citep{DBLP:journals/corr/BahdanauCB14}. 
Transformers have achieved great success in various natural language processing tasks.
Nevertheless, the quadratic computation complexity can be costly.
Some works try to address the limitations of self-attention, including Reformer, Sparse Transformer, Longformer, Linformer \citep{child2019sparsetransformer,DBLP:conf/iclr/KitaevKL20,DBLP:journals/corr/abs-2006-04768}, etc.
They successfully reduce the complexity of self-attention and thus enable processing longer sequences.
However, most of them still require linear memory space complexity.

Transformer-XL \citep{DBLP:conf/acl/DaiYYCLS19} re-introduces the concept of memory and recurrence.
It caches each layer's hidden states of self-attention into a fixed-size queue and re-uses them in the later attention computation.
However, the memory as raw hidden states cannot effectively compress high-level information.
Thus, Transformer-XL in practice needs a massive memory size to perform well, and spends huge computation in using its memory.
Compressive Transformer \citep{DBLP:conf/iclr/RaePJHL20} improves upon Transformer-XL by further compressing its memories into fewer vectors via a compression network.
However, as mentioned in the papers, both Transformer-XL and Compressive Transformer discard the information from the distant past, which causes a theoretical maximum temporal range given the fixed memory size.

Inspired by the previous external memory networks, we propose Memformer, which incorporates a fixed-size external dynamic memory combined with the recent Transformer architecture.
Memformer interacts with its external dynamic memory through the memory reading and writing modules.
Also, we introduce a forgetting mechanism to improve the effectiveness of memorizing new information.
By utilizing recurrence and a fixed-size memory, our model has a theoretically infinite temporal range of memorization and implies a linear computation complexity and constant memory space complexity.
As the traditional back-propagation through time (BPTT) has an unaffordable memory cost in our model, we introduce a new optimization scheme, memory replay back-propagation (MRBP), to significantly reduce the memory cost in training recurrent neural networks with large size of memory representations.

We evaluate Memformer on the autoregressive image generation and language modeling task. Experimental results show that Memformer performs on par with Transformer and Transformer XL with large memory size, while being much more efficient in terms of computation speed and memory space consumption. 
We also conduct an analysis showing that Memformer can retain information for an extended period.

%% file: related_work.tex
\section{Related Work}

This section introduces some recent research directions that aim to alleviate the quadratic cost of self-attention.
Moreover, we analyze their assumptions and limitations under the autoregressive setting to provide a broader view of these models.

\subsection{Sparse Attention}


One influential direction is to replace the full self-attention with sparse attention patterns to speed up the computation.
\citet{child2019sparsetransformer} proposed Sparse Transformer, using a block sparse attention pattern to reduce the computation complexity to $\mathcal{O}(N \sqrt{N})$.
Later, Longformer \citep{DBLP:journals/corr/abs-2004-05150} and Big Bird \citep{zaheer2020bigbird} further explored this direction and proposed an even more sparse attention pattern to reduce the cost to $\mathcal{O}(N)$.
They introduced global tokens to encode the information from the entire sequence and kept the self-attention to the closest $k$ tokens and the global tokens to achieve linear complexity.
Although linear sparse attention's theoretical soundness is proven for bidirectional encoders, it does not hold for the decoder.
The main reason is that the global tokens cannot leak information to the future tokens in the autoregressive setting, where all the tokens can only see their previous tokens.
Thus, linear sparse attention cannot guarantee a token to see its all past tokens.
Only Sparse Transformer here with $\mathcal{O}(N \sqrt{N})$ complexity can theoretically cover all the past tokens for the sequence generation.

\subsection{Linear Attention}

Another direction is focusing on improving the softmax operation in the self-attention.
Linformer \citep{DBLP:journals/corr/abs-2006-04768} reduced the complexity to $\mathcal{O}(N)$ by projecting the entire sequence to a constant size of keys and values, but this method has not been applied to autoregressive decoding.
Performer \citep{DBLP:journals/corr/abs-2009-14794} and Linear Transformer \citep{DBLP:conf/icml/KatharopoulosV020} used a linear dot-product of kernel feature maps to replace softmax.
However, for Linear Transformer under the autoregressive setting, it needs to compute the cumulative summation to aggregate the history information.
This assumption is too strong if the input sequence is long and the length is not fixed. 
After thousands of steps, the numerical values can become very large due to the summation, causing overflow and gradient instability.

\subsection{Recurrence and Memory}

Applying recurrence and memory to Transformers is an orthogonal direction comparing to the efficient attention approaches.
If the memory size is constant, recurrence enables the model to have constant memory complexity during inference.
There are mainly two works exploring this direction.
Transformer-XL \citep{DBLP:conf/acl/DaiYYCLS19} used relative positional encoding and consisted of a segment-level recurrence mechanism to encode beyond a fixed-length context.
Compressive Transformer \citep{DBLP:conf/iclr/RaePJHL20} extended from Transformer-XL by further compressing the previous cached hidden states to achieve a longer context.
However, using past hidden states as memory would cause a theoretical maximum temporal range of context, meaning that a token is not guaranteed to see all the past tokens.
Thus, in practice, Transformer-XL and Compressive Transformer need huge memory size to achieve good performance.

\subsubsection{Dynamic Memorization}

Within the scope of memory networks, there are dynamic memorization techniques.
Different from Transformer-XL which stores the token-level history representations as memory, dynamic memorization does not have a theoretical upper bound for the temporal range.
Neural Turing Machine (NTM) \citep{DBLP:journals/corr/GravesWD14} and Differential Neural Computer (DNC) \citep{DBLP:journals/nature/GravesWRHDGCGRA16} are two early models that can control external memory resources to achieve long-lasting memory.
However, their complex memory mechanisms cause them to be slow and unstable during training.
In this work, we propose a dynamic memorization mechanism to achieve more efficient memory representations.

%% file: methods.tex
\SetKwComment{Comment}{$\triangleright$\ }{}

\section{Methods}

\begin{figure}[t]
    \raggedright
    \includegraphics[width=0.44\textwidth]{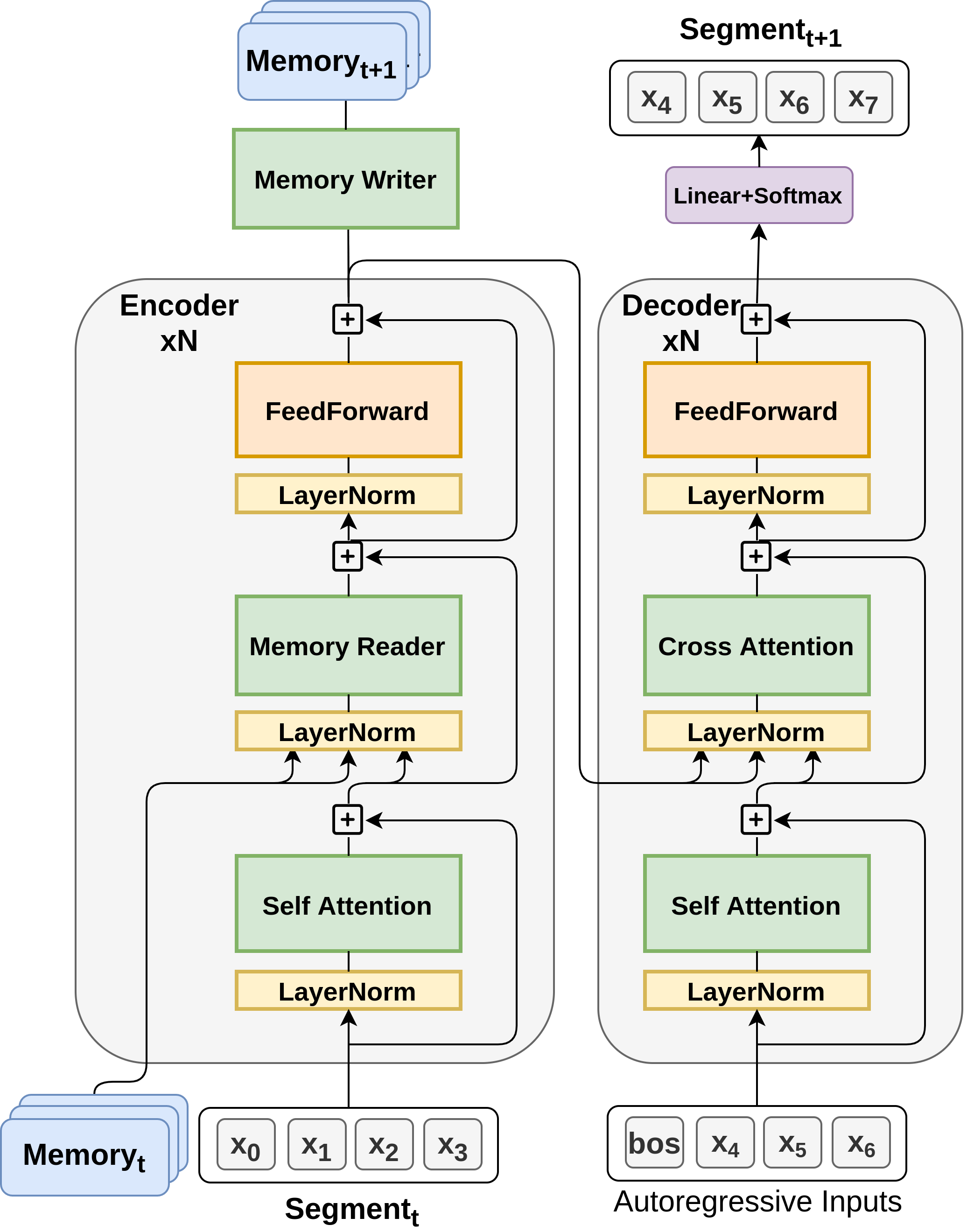}
    \caption{Memformer overall architecture for the encoder (left) and decoder (right). Transformer encoder is responsible to interact with the memory. Sequence modeling is achieved by predicting the next segment conditioned to the current segment and memory.}
    \label{fig:overall_arch}
\end{figure}

In this section, we first formalize the segment-level sequence modeling.
Then, we present the memory reading and writing modules.
Finally, we explain the memory replay back-propagation (MRBP) algorithm used for training.

\subsection{Segment-level Sequence Modeling}

Given a sequence of $N$ tokens $x_1, x_2, \ldots, x_N$, an
standard language model learns the joint probability of the sequence by taking the product of each token's probability conditioned to the previous tokens,
which is defined as:
\begin{equation*}
    P(x) = \prod_t P(x_t| x_{<t})
\end{equation*}

When we have a large external memory system to store the history information, we cannot afford to interact with memory for every token.
The workaround is to process a long sequence at the segment level.
We can split a sequence into $T$ segments and each segment has $L$ tokens: $s_t=\{x_{t,1}, x_{t,2}, \ldots x_{t,L}\}$.

Because a bidirectional encoder is better at extracting word representations, we apply a Transformer encoder-decoder here.
The encoder's role is to encode the segment $s_t$ and inject the information into the memory $M_t$, while it also retrieves past information from the previous timestep's memory $M_{t-1}$.
The encoder's final output will be fed into the decoder's cross attention layers to predict the token probabilities of the next timestep's segment $s_{t+1}$ with standard language modeling.
\begin{align*}
    M_t    & = \text{Encoder}(s_t, M_{t-1})                                        \\
    P(s_t |s_{<t} ) & = \prod_{n=1:L} P_{\text{Decoder}}(x_{t,n} \, | \, x_{t,<n}, M_{t-1}) \\
    P(x)   & = \prod_{t=1:T} P_{\text{Model}}(s_{t}|s_{<t})
\end{align*}

At each timestep, given a segment as the input, the model needs to continue that segment by generating the next text segment, and the generated segment will be fed back into the model again.
Since the memory stores all the past information, we can autoregressively generate all the token segments in a sequence. In this fashion, we can model the entire long sequence.

Figure~\ref{fig:overall_arch} shows the overall architecture of Memformer.
We will further explain each component and the implementation in the following sections.

\subsection{External Dynamic Memory Slots}

External dynamic memory (EDM) is a data structure that stores high-level representations of past inputs.
“Dynamic'' means that the model interactively encodes and retrieves the information from memory in a recurrent manner.
This contrasts with static memory design, where the memory is stored statically and does not change during the inference.

In our design, we allocate a constant $k$ number of vectors as the external dynamic memory.
At each timestep $t$, we can have $M_t = [m_t^0, m_t^0, \ldots, m_t^k]$.
For each sample in the batch, they have separate memory representations.
Therefore, similar to RNN during inference, the memory consumption will be constant no matter how long the input sequence is.
We name it memory slots because each slot is working individually to have different representations.
The following sections will explain how the model manages to read and write this memory.

\subsection{Memory Reading}

\begin{figure}[t]
    \centering
    \includegraphics[width=0.40\textwidth]{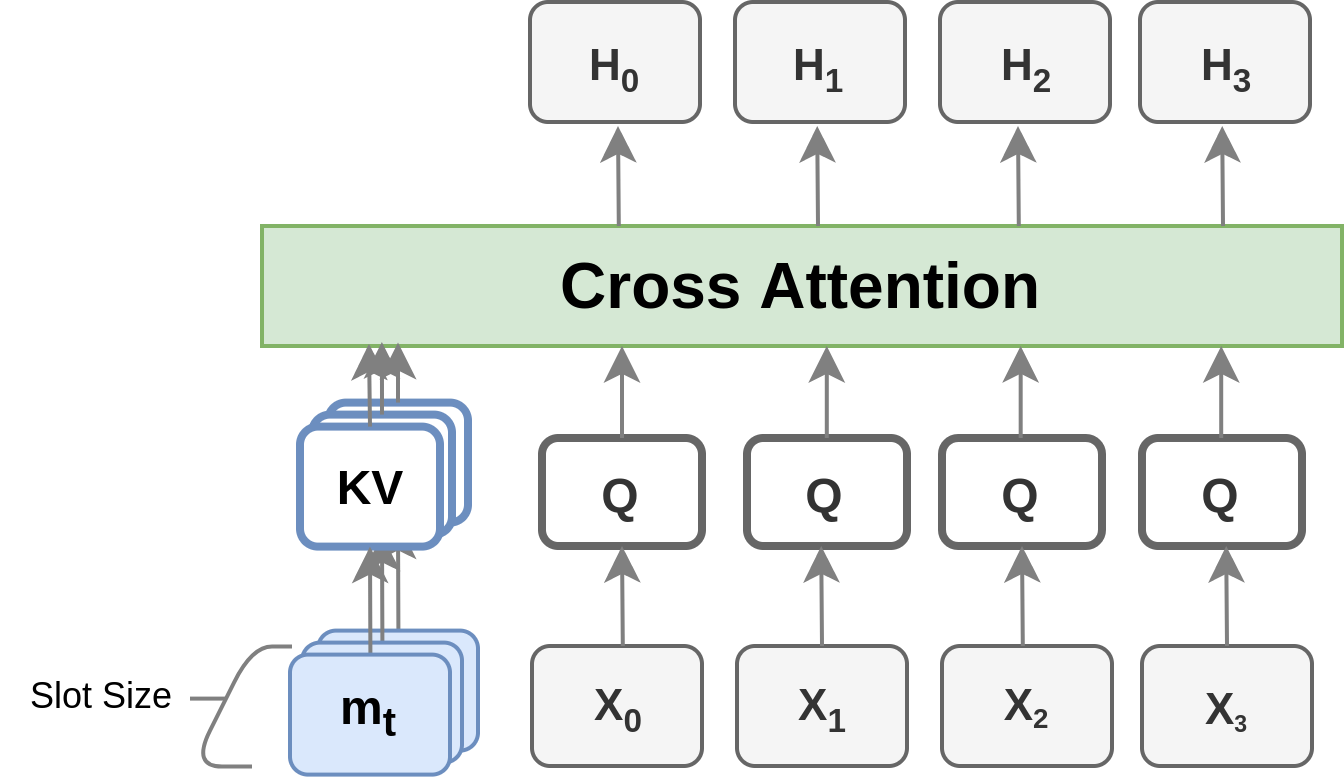}
    \caption{Memory Reading. The input sequence $x$ attends over all the memory slots to retrieve the history information. }
    \label{fig:memread}
\end{figure}

For each input segment sequence, the model needs to read the memory to retrieve relevant past information.
We leverage the cross attention to achieve this function:
\begin{align}
    Q_x, K_M, V_M & = x W_Q, M_t W_K, M_t W_V         \\
    A_{x, M}      & = \text{MHAttn} (Q_x, K_M)        \\
    H_x           & = \text{Softmax}(A_{x, M}) \, V_M
\end{align}

MHAttn refers to Multi-Head Attention.
Memory slot vectors are projected into keys and values, and the input sequence $x$ is projected into queries.
Then the input sequence's queries attend over all the memory slots' key-value pairs to output the final hidden states.
This enables the model to learn the complex association of the memory. Figure~\ref{fig:memread} shows the illustration.

Memory reading occurs multiple times as every encoder layer incorporates a memory reading module.
This process ensures a higher chance of successfully retrieving the necessary information from a large memory.

\subsection{Memory Writing}

Memory writing involves a slot attention module to update memory information and a forgetting method to clean up unimportant memory information.
Contrary to memory reading, memory writing only happens at the last layer of the encoder. 
This helps to store the high-level contextual representations into the memory.
In practice, we append some classification tokens to the input sequence to better extract the sequence representations. 

\begin{figure}[h]
    \centering
    \includegraphics[width=0.44\textwidth]{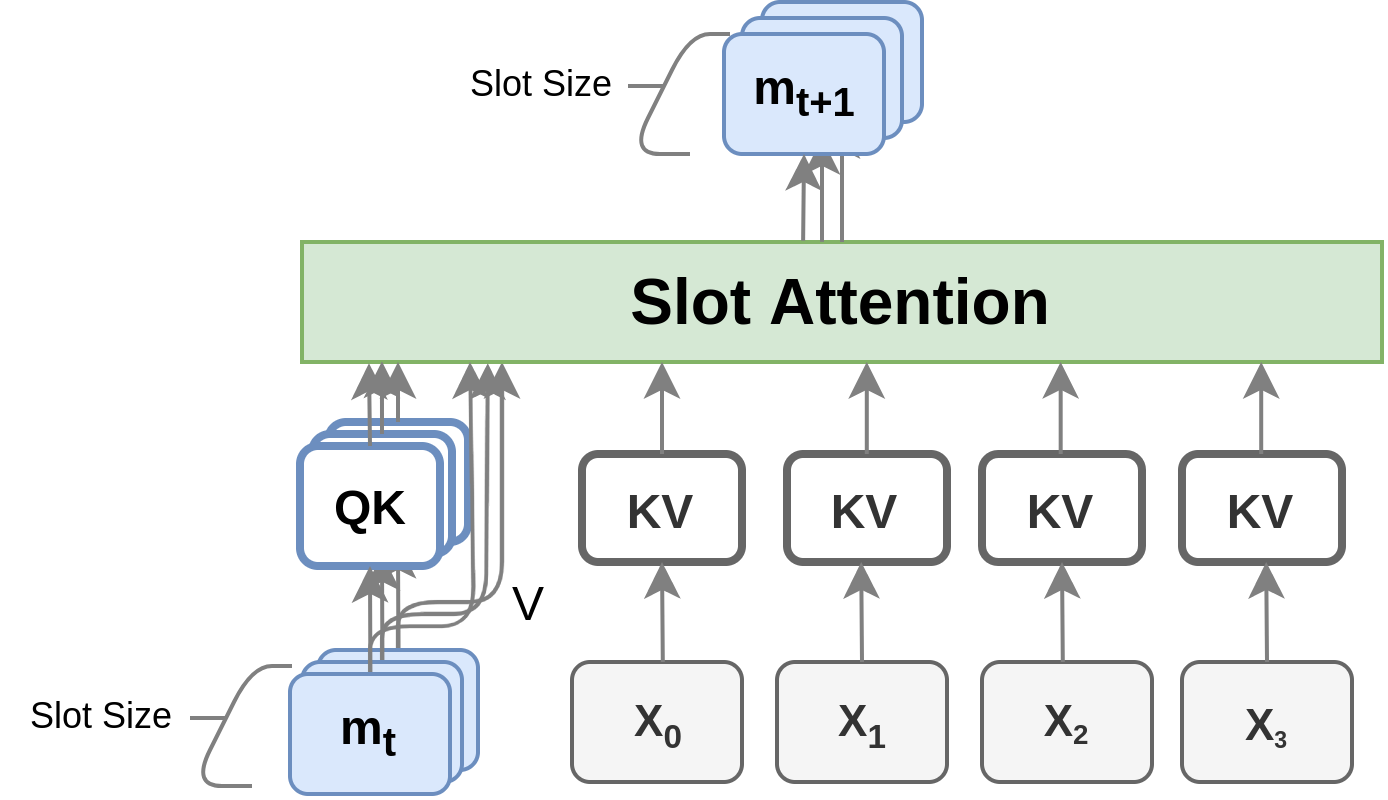}
    \caption{Memory Writing. Each memory slot attends over itself and the input sequence representations to produce the next timestep's memory slot.}
    \label{fig:memwrite}
\end{figure}

\subsubsection{Update via Memory Slot Attention}

Figure~\ref{fig:memwrite} shows how memory is updated with the current segment's information.
Each slot is separately projected into queries and keys.
The segment token representations are projected into keys and values.
Slot attention means that each memory slot can only attend to itself and the token representations.
Thus, each memory slot cannot write its own information to other slots directly, as memory slots should not be interfering with each other.

\begin{align}
    Q_{m^i},  K_{m^i} &= m^i W_Q, m^i W_K                         \\
              K_x, V_x &= x W_K, x W_V                            \\
             A'_{m^i} =&  \text{MHAttn} (Q_{m^i}, [K_{m^i}; K_x])
\end{align}

When we compute the final attention scores, we divide the raw attention logits with a temperature $\tau$ ($\tau < 1$).
This operation sharpens the attention distribution, which makes the writing focusing on fewer slots or token outputs.
\begin{equation}
    A_{m^i} = \frac{exp(A_i' / \tau)}{\sum_j exp(A_j' / \tau)}
\end{equation}
Finally, the next timestep's memory is collected with by attention.
\begin{equation}
    {m^i_{t+1}}' = \text{Softmax}(A_{x, M}) \,  [m_t^i; V_x]
\end{equation}
The attention mechanism helps each memory slot to choose to whether preserve its old information or update with the new information.

\subsubsection{Implementation of Memory Writer}


Since each memory slot stores the information independently, we design a special type of sparse attention pattern.
Each slot in the memory can only attend over itself and the encoder outputs.
It aims to preserve the information in each slot longer over the time horizon.
When a slot only attends itself during writing, the information will not be changed in the next timestep.

\subsubsection{Forgetting Mechanism}
Forgetting is crucial for learning as it helps to filter out trivial and temporary information to memorize more important information.
LSTM introduces the forget gate \citep{DBLP:journals/neco/GersSC00} to reset its memory state,
and the forget gate is proven to be the most important component in the LSTM \citep{DBLP:journals/corr/abs-1804-04849}.

In this work, we introduce a forgetting mechanism called \textit{Biased Memory Normalization} (BMN), specifically designed for our slot memory representations.
We normalize the memory slots for every step to prevent memory weights from growing infinitely and maintain gradient stability over long timesteps.
To help forget the previous information, we add a learnable vector $v_{\text{bias}}$ to it.
Also, naturally the initial state $v^i_{\text{bias}}$ is  after normalization.
\begin{align*}
    m^i_{t+1} & \leftarrow {m^i_{t+1}}+ v^i_{\text{bias}} \\
    m^i_{t+1} & \leftarrow \frac{m^i_{t+1}}{||m^i_{t+1}||} \\
    m^i_{0} & \leftarrow \frac{v^i_{\text{bias}}}{||v^i_{\text{bias}}||}
\end{align*}

\begin{figure}[t]
    \centering
    \includegraphics[width=0.35\textwidth]{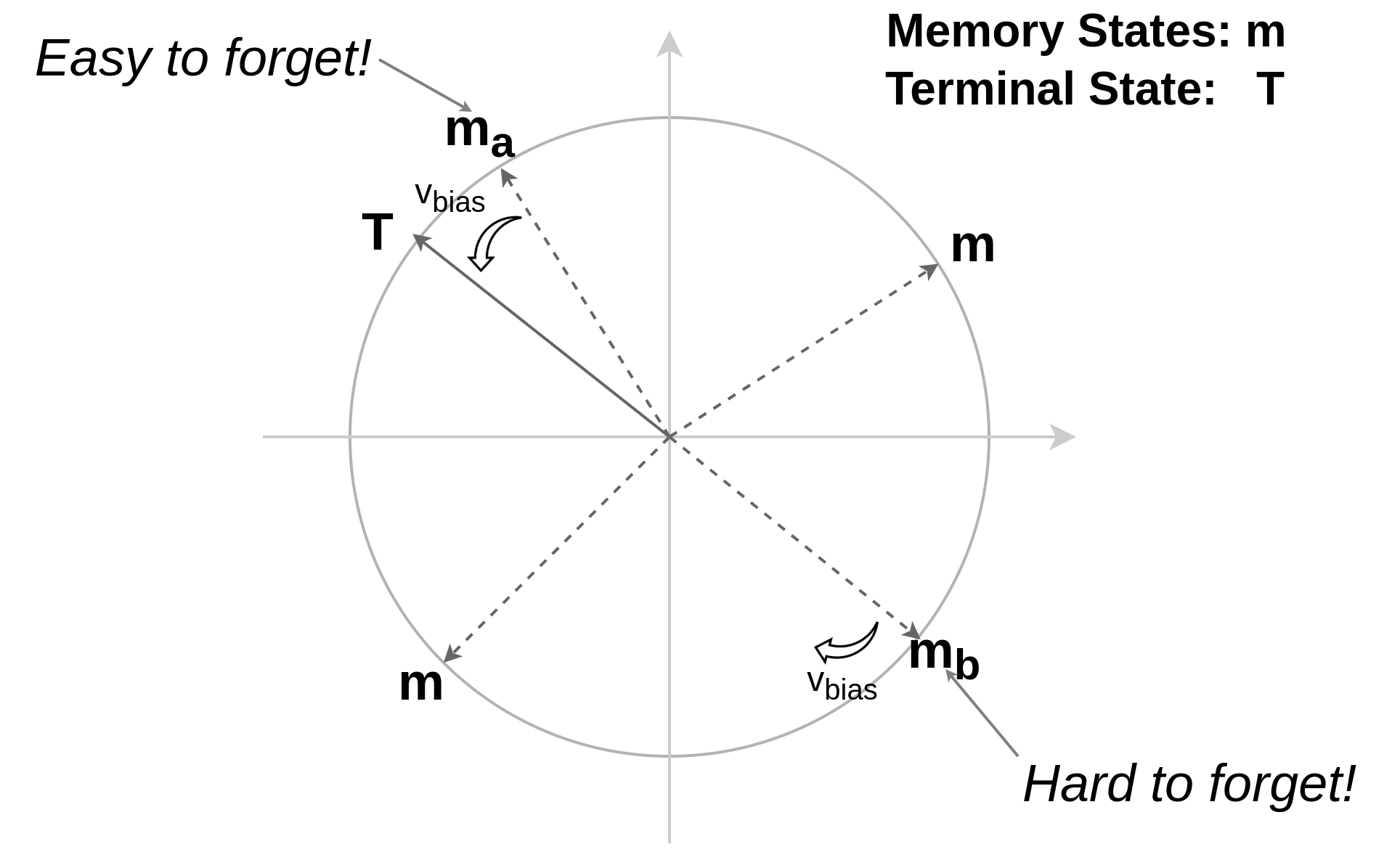}
    \caption{Illustration of forgetting. Memory slot $m_a$ is easy to be forgotten, while $m_b$ is hard to be forgotten.}
    \label{fig:bmn}
\end{figure}

In Figure~\ref{fig:bmn}, we illustrate the forgetting mechanism with the learnable bias vector $v_{\text{bias}}$.
Because of the normalization, all memory slots will be projected onto a sphere distribution.
Here, we demonstrate with a 2D sphere for simplicity.

$v_{\text{bias}}$ here controls the speed and the direction of forgetting.
When adding $v_{\text{bias}}$ to the memory slot, it would cause the memory to move along the sphere and forget part of its information.
If a memory slot is not updated for many timesteps, it will eventually reach the terminal state $T$ unless the new information is injected.
The terminal state is also the initial state, and it is learnable.

The speed of forgetting is controlled by the magnitude of $v_{\text{bias}}$ and the cosine distance between $m'_{t+1}$ and $v_{\text{bias}}$.
For example, $m_b$ is nearly opposite to the terminal state, and thus would be hard to forget its information.
$m_a$ is closer to the terminal state and thus easier to forget.

\LinesNumbered
\begin{algorithm}[t]
    \caption{Memformer Update}
    \label{algo:mrbp}
    \KwIn{rollout=[$x_t, x_{t+1}, \ldots, x_T$]: \texttt{a list containing previous inputs} \newline
    memories=[$M_t, M_{t+1}, \ldots, M_T$]: \texttt{memory from the previous}
    }
    \Comment{Initialize a list for back-propagation}
    replayBuffer = [$M_t$] \\
    \Comment{Forward pass \& no gradient}
    \For {$t=t, t+1, \ldots, T - 1$}
    {
        $M_{t+1}$, $\_$ = Model($x_t$, $M_t$) \\
        replayBuffer.append($M_{t+1}$)
    }
    \Comment{Backward pass with gradient}
    $\nabla M_{t+1} = 0$ \\
    \For {$t=T, T-1, \ldots, t+1, t$}
    {   \Comment{Recompute}
        $M_{t+1}$, $O_{t}$ = Model($x_t$, $M_t$) \\
        $loss$ = $f_{loss}(O_{t})$ \\
        $loss$.backward() \\
        $M_{t+1}$.backward($\nabla M_{t+1}$) \\
        $\nabla M_{t+1}$ = $\nabla M_{t}$
    }
    \Comment{Update and pop the oldest memories}
    memories = replayBuffer \\
    memories.pop() \\
\end{algorithm}

\subsection{Memory Replay Back-Propagation}

Memformer relies on the external memory to process a sequence.
At inference time, there is no additional memory cost because of the fixed-size memory design.
Nevertheless, during training, it would require back-propagation through time (BPTT) so that the memory writer network can be trained to retain long-term information.
The problem with traditional BPTT is that it unrolls the entire computational graph during the forward pass and stores all the intermediate activations.
This process would lead to impractically huge memory consumption for Memformer.

A favorable existing approach to eliminate this problem is gradient checkpointing  \citep{DBLP:journals/corr/ChenXZG16}.
The algorithm can significantly reduce the memory cost of a large neural network.
However, the standard gradient checkpointing still needs to compute all the nodes in the computational graph and store unnecessary activations during the forward pass.
We propose Memory Replay Back-Propagation (MRBP), a more efficient variant of gradient checkpointing, by replaying the memory at each timestep to accomplish gradient back-propagation over long unrolls.

The algorithm takes an input with a rollout $x_t, x_{t+1}, \ldots, x_T$ and the previous memories $M_t, M_{t+1}, \ldots, M_T$ if already being computed.
MRBP only traverses the critical path in the computational graph during the forward pass and recomputes the partial computational graph for the local timestep during the backward pass.
It then obtains each timestep's memory and stores those memories in the replay buffer.
The full algorithm is described in Algorithm~\ref{algo:mrbp}.
The experiments of memory cost reduction with MRBP is in the Appendix A.

%% file: experiments.tex
\section{Experiments}


\subsection{Computation and Memory Cost}

\begin{figure*}[h]
    \centering
    \begin{subfigure}[b]{0.44\textwidth}
        \includegraphics[width=\textwidth]{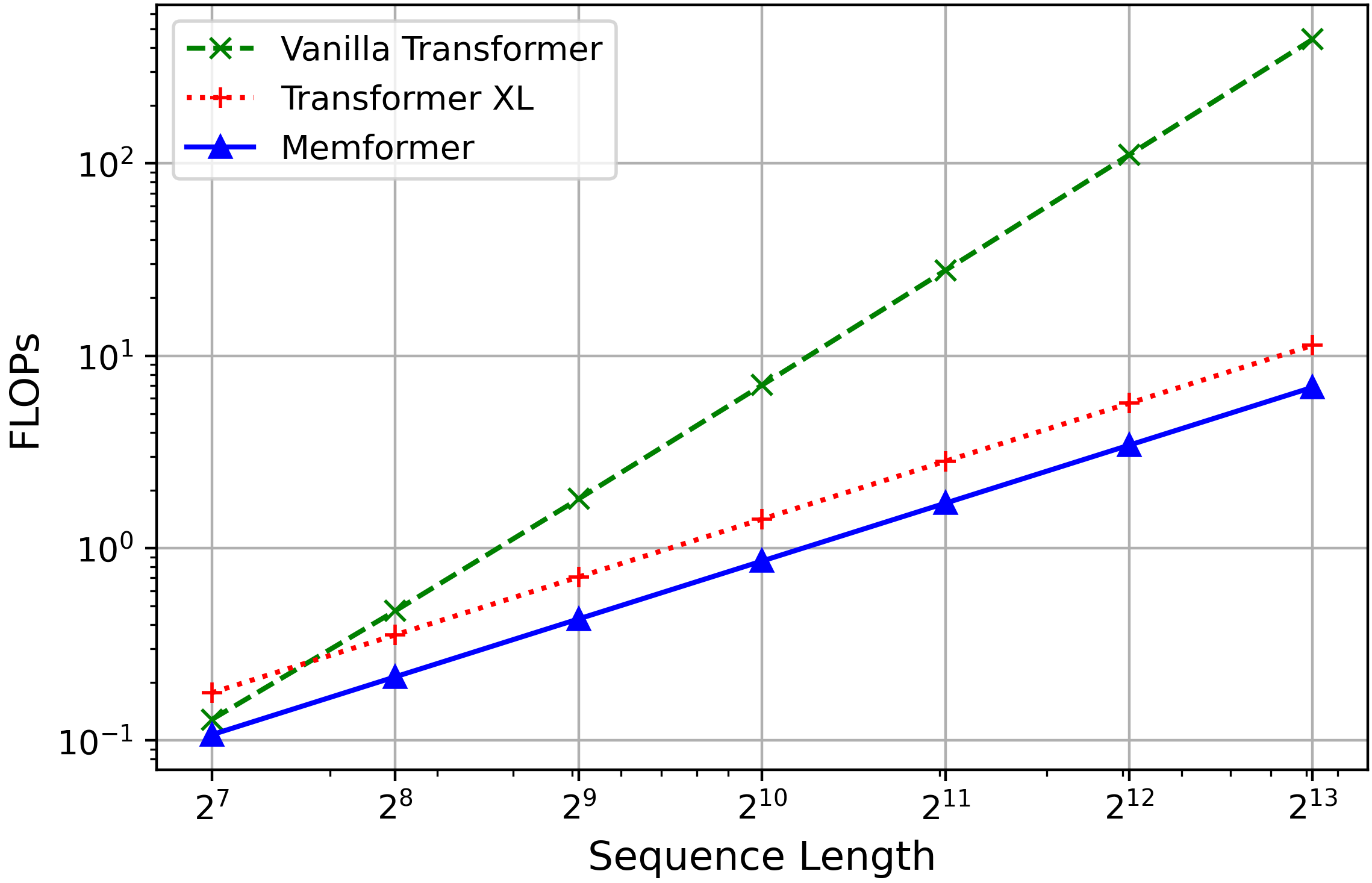}
    \end{subfigure}
    \hspace{1em}
    \centering
    \begin{subfigure}[b]{0.44\textwidth}
        \includegraphics[width=\textwidth]{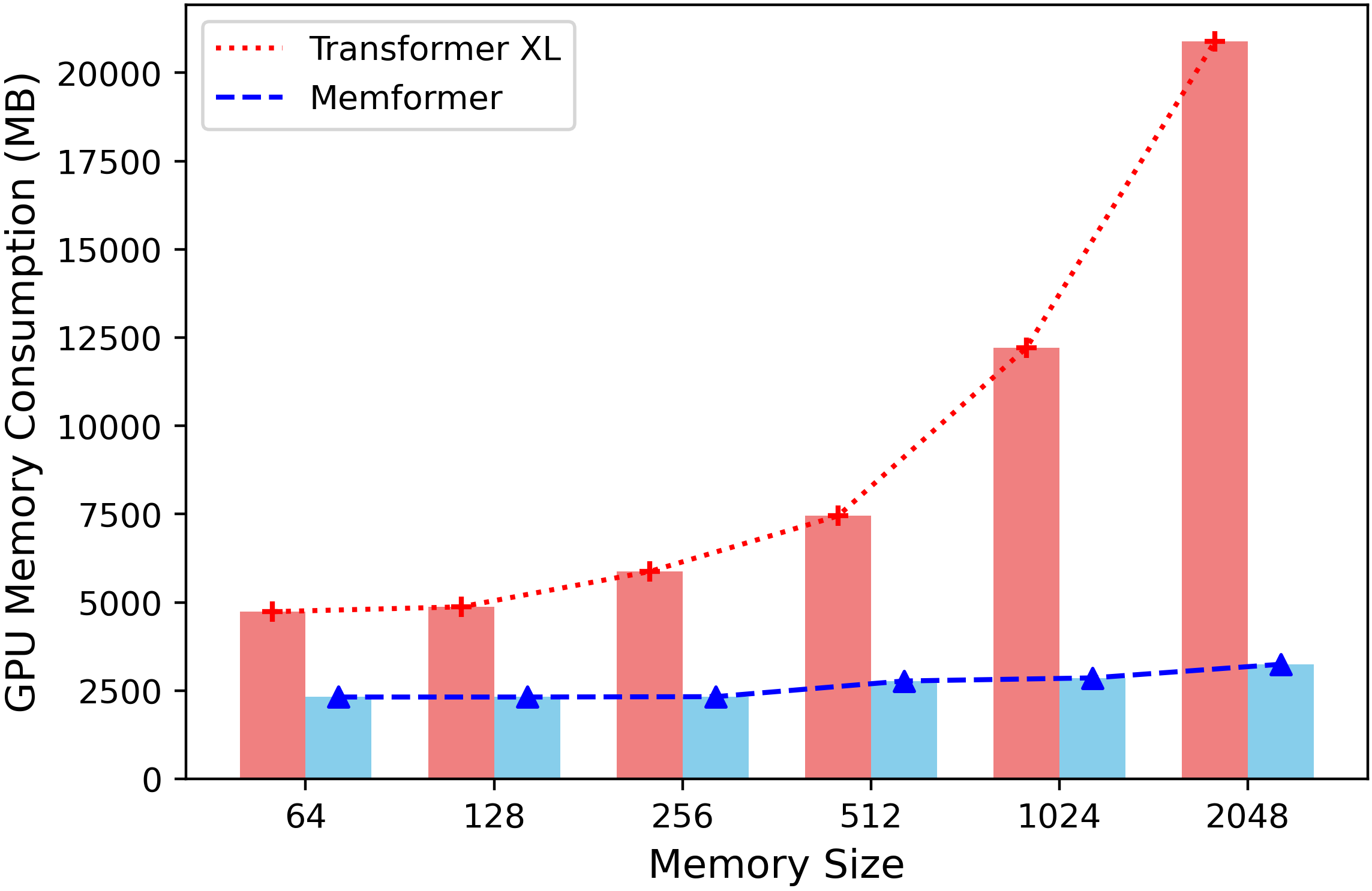}
    \end{subfigure}
    \caption{Comparison of the number of FLOPs and GPU memory consumption for Vanilla Transformer, Transformer-XL, and Memformer.}
    \label{fig:compution_cost}
\end{figure*}

We experimented the computation and memory cost of Vanilla Transformer, Transformer-XL, and Memformer.
For Vanilla Transformer, it has to increase the input sequence length to encode more tokens.
Its cost is $O(N^2)$ where $N$ is the sequence length.
Transformer-XL and Memformer use memory to store the history information, and the input sequence length is a constant value.
Thus, their computation complexity is $O(N)$.

As a trade-off, for both Transformer-XL and Memformer, the memory size is then an important factor to affect the capacity of storing the history information.
Transformer-XL stores the past hidden states for all layers as memory.
If $L$ is the number of layers, and $K$ is the memory size, then the memory cost is $O(K \times L)$.
Memformer only stores $K$ vectors as memory with cost $O(K)$.

To better illustrate the difference, Figure~\ref{fig:compution_cost} shows the number of FLOPs (floating-point operations) versus sequence length (left) and the GPU memory consumption versus memory size on the actual models (right).
The sequence length is increased from $128$ to $8,192$.
Here, Memformer and Transformer-XL had the same number of parameters.
From the figure, Vanilla Transformer has the largest computation cost growth.
Memformer's costs grew linearly with the sequence length and achieved better efficiency than Transformer-XL.
Then, we compared the GPU memory consumption.
We tested the memory size ranging from $64$ to $2,048$, with a batch size $16$ for better visibility of memory cost difference.
Transformer-XL's memory consumption grew rapidly with the memory size, while Memformer is more efficient with large memory size.
In large memory size setting, Memformer uses 8.1x less memory space.

\subsection{Autoregressive Image Generation}

\begin{table}[ht]
    \centering
    \resizebox{0.47\textwidth}{!}{
        \begin{tabular}{l|cc}
            \toprule
            Model                     & \#FLOPs (B)       & Perplexity $\downarrow$     \\
            \midrule
            LSTM                      & $52.5$         & $1.698$          \\
            Transformer Decoder       & $41.3$         & $1.569$          \\
            Transformer-XL            &               &                \\
            \quad memory=56      & $5.6$          & $1.650$          \\
            \quad memory=224     & $15.6$         & $1.618$          \\
            \quad memory=784     & $49.1$         & $1.611$          \\
            \midrule
            Memformer                 &               &                \\
            \quad 4 encoder+8 decoder & $\mathbf{5.0}$ & $\mathbf{1.555}$ \\
            \midrule
            Memformer Ablation        &               &                \\
            \quad 2 encoder+6 decoder &               &                \\
            \quad \quad memory=64     & $3.9$          & $1.594$          \\
            \quad \quad memory=32     & $3.9$          & $1.600$          \\
            \quad \quad memory=16     & $3.9$          & $1.604$          \\
            \quad \quad memory=1      & $3.9$          & $1.627$          \\
            \quad 4 encoder+4 decoder & $3.6$          & $1.628$          \\
            \quad w/o memory          & $1.8$          & $1.745$          \\
            \quad temperature=1.0     & $3.9$          & $1.612$          \\
            \quad w/o forgetting      & $3.9$          & $1.630$          \\
            \quad w/o multi-head      & $3.9$          & $1.626$          \\
            \bottomrule
        \end{tabular}
    }
    \caption{Results for autoregressive image generation. Our method only takes about $10\%$ FLOPs of the best Transformer-XL model.}
    \label{tab:mnist}
\end{table}

Recent research \cite{ramesh2021zeroshot} demonstrates the approach of treating an image as a long sequence for image generation.
Thus, we evaluated our model on the MNIST \citep{lecun-mnisthandwrittendigit-2010} image generation task with sequence modeling.
Each image of size $28\times28$ was reshaped into a sequence of $784$ tokens, and
the $8$-bit gray-scale was turned to a $256$ vocabulary size.

For the baselines, LSTM had $4$ layers and $512$ hidden size.
Transformer Decoder had $8$ layers and could take all the $784$ tokens as the input.
Transformer-XL had 8 layers.
All the models had the same $128$ hidden size, $4$ attention heads, $32$ head size, and $256$ feedforward size.
Memformer was tested with default memory size $64$. The default memory writer temperature was set to $0.25$.
We also conducted ablation studies to examine the contribution of various components.

\begin{table}[h]
    \centering
    \resizebox{0.482\textwidth}{!}{
        \begin{tabular}{l|cc}
            \toprule
            Model                                             & \#FLOPs (B) & PPL $\downarrow$ \\
            \midrule
            Transformer-XL base                &         &                  \\
            \quad memory=1600                                 & $250$    & $23.95$            \\
            \quad memory=1024                                 & $168$    & $23.67$            \\
            \quad memory=512                                  & $94$     & $23.94$            \\
            \quad memory=256                                  & $58$     & $25.39$            \\
            \quad memory=128                                  & $39$     & $25.60$            \\
            \quad memory=32                                   & $26$     & $27.22$            \\
            Compressive Transformer  &   & \\
            \quad memory= 512 compress=512 & 172 &  23.23 \\
            \midrule
            Memformer                                 &         &                  \\
            \quad 4 encoder + 16 decoder                      & $\mathbf{54}$     & $\mathbf{22.74}$            \\
            \midrule
            Memformer Ablation                                &         &                  \\
            \quad 4 encoder + 12 decoder                      & $48$     & $23.91$            \\
            \quad memory=512                                  & $35$     & $23.30$            \\
            \quad w/o memory                                  & $31$     & $25.57$            \\
            \bottomrule
        \end{tabular}
    }
    \caption{Experimental results on language modeling. Our method is $3.2$ times faster here.}
    \label{tab:wiki103_results}
\end{table}

Table~\ref{tab:mnist} shows the experimental results.
We report median from three trials.
Our Memformer with $4$ layers of encoder and $8$ layers of decoder achieved the best performance ($1.555$),
while only using nearly $10\%$ of FLOPs compared to the best Transformer XL baseline with memory size of $784$ ($1.611$).
Its performance was even better than the Transformer Decoder with the entire input sequence.
We hypothesized that this observation was due to the extra parameters from the 4 layers of encoder.
Therefore, we conducted an ablation study by having various numbers of encoder and decoder layers.
If we reduce the number of decoder layers in Memformer ($4$ encoder+$4$ decoder), the performance dropped as shown ($1.628$).
Results indicated that the number of decoder layers was important for the performance.
Overall, Memformer outperformed Transformer-XL with a much lower computation cost.

The performance increased as the memory size increased.
Moreover, when we completely removed the memory, Memformer performed terribly, signifying the importance of the encoded information in the memory.
Other components such as forgetting mechanism, memory writer temperature, multi-head attention were proven to contribute to the final performance as well.

\subsection{Language Modeling}

We also conducted experiments on WikiText-103 \citep{DBLP:conf/iclr/MerityX0S17}, which is a long-range language modeling benchmark.
It contains $28K$ articles with an average length of $3.6K$ tokens per article.
Due to the limitation of computational resources, we are unable to experiment on the more recent PG19 \cite{DBLP:conf/iclr/RaePJHL20} dataset.
To study the computation cost and memory efficiency, we test with Transformer-XL base with $16$ layers, $512$ hidden size, $2,048$ feedforward size, $64$ head size, and $8$ heads. The details are in the Appendix.

Memformer has the same hidden size, feedforward size, head size, and number of heads.
We also re-implement a version of Compressive Transformer of the same size as there is no official implementation.
The memory length is set to 512, and the compressive memory length is 512.
The compression ratio is 4.
The target sequence length for all models was set to $128$.
We test the performance under various memory sizes.

Table~\ref{tab:wiki103_results} summarizes the results on WikiText-103 test set.
We report the number of inference FLOPs (billions) and perplexity median from three trials.
As Transformer-XL's memory size increased, the perplexity dropped as expected, but the the number of FLOPs grew quickly because the attention length was also increased.
The perplexity stopped decreasing after we increased the memory size to $1,600$.
We suspect that since the average number of tokens in WikiText-103 is $3,600$, a larger memory size would bring noises and hence did not further improve the performance compared to a smaller memory size ($1,024$).
Compressive Transformer achieves slightly better performance with extra FLOPS compared to Transformer XL with memory size 1024.

Memformer with $4$ encoders, $16$ decoders, and $1,024$ memory size achieved the best performance.
It required much less computation cost ($54$) and performed much better than Transformer-XL with $1,024$ memory size, supporting that Memformer has a more efficient memory representation.

In the ablation studies, to compensate for the extra number of encoder layers, we reduced the number of decoder layers to $12$.
The final performance was close to Transformer-XL, but Memformer used a much smaller number of FLOPs.
Also, memory size was important for Memformer, as the performance dropped after the memory size is reduced to $512$.
When we completely removed the memory module by removing the memory writer and memory reading cross attention, the perplexity increased to $25.57$, which is similar to Transformer-XL with a memory size of $128$.

\subsubsection{Memory Writer Analysis}

\begin{figure}[ht]
    \centering
    \includegraphics[width=0.4\textwidth]{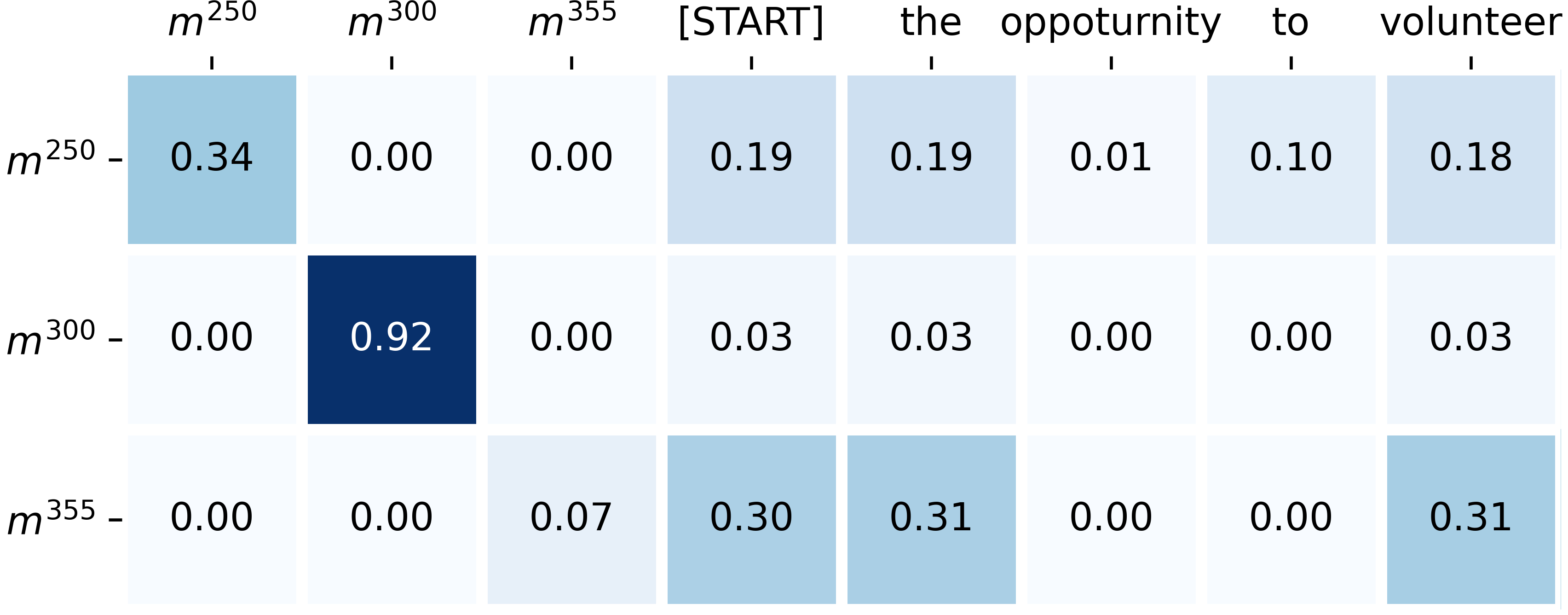}
    \caption{Visualization of three types of memory slots.}
    \label{fig:memattn}
\end{figure}

It is interesting to interpret how memory writer updates the memory slots.
We analyzed the attention outputs from the memory writer.
We roughly categorized the memory slots into three different types and visualized three examples with normalized attention values in Figure~\ref{fig:memattn}.

We picked the memory slot $m^{250}$, $m^{300}$ , and $m^{355}$.
During the middle of processing a document, around $60\%$ to $80\%$ of the memory slots are like $m^{300}$.
Their attention focused on themselves, meaning that they were not updating for the current timestep.
This suggests that the memory slots can carry information from the distant past.

For the second type, the memory slot $m^{250}$ had some partial attention over itself and the rest of attention over other tokens.
This type of memory slots is transformed from the first type of memory slots, and at the current timestep they aggregate information from other tokens.

The third type of memory slot looks like $m^{355}$. It completely attended to the input tokens. At the beginning, nearly all memory slots belong to this type, but later only $5\%$ to $10\%$ of the total memory slots account for this type.
We also found that the forgetting vector's bias for $m^{355}$ had a larger magnitude ($3.20$) compared to some other slots ($1.15$), suggesting that the information was changing rapidly for this memory slot.

\begin{figure}[h]
    \centering
    \includegraphics[width=0.482\textwidth]{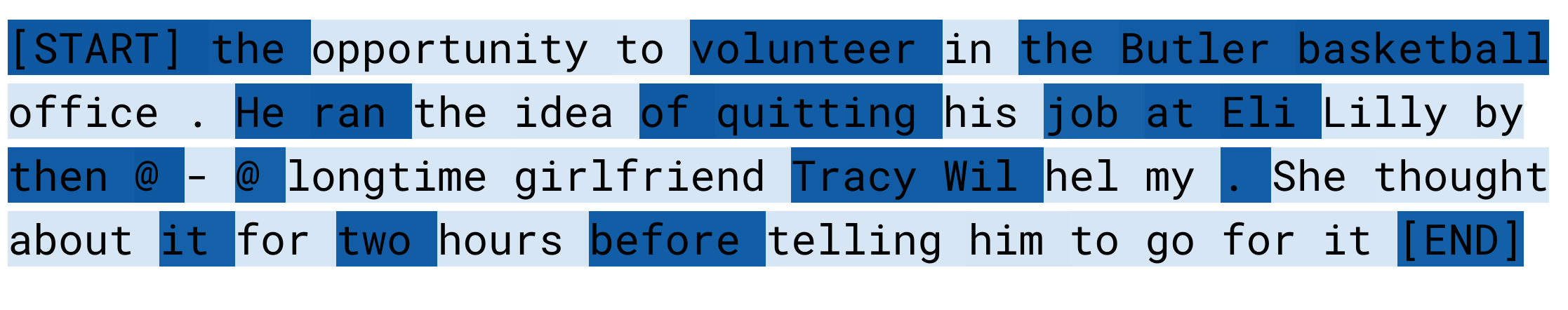}
    \caption{Visualization of the memory writer's attention.}
    \label{fig:wordattn}
\end{figure}

To better understand how the slot $m^{355}$ update its information, we visualized its attention on an example input sequence in Figure~\ref{fig:wordattn}.
It shows that this slot learned a compressed representation of the sentence by attending over some named entities and verbs, which is consistent with human cognition.

%% file: conclusion.tex
\section{Conclusion}


We presented Memformer, an autoregressive model which utilizes an external dynamic memory to efficiently process long sequences with a linear time complexity and constant memory complexity.
Along with Memformer, we introduced a new optimization scheme, Memory Replay Back-propagation, which enables training recurrent neural networks with large memory.
Experimental results showed that Memformer achieved comparable performance with great efficiency, and was able to preserve information from the distant past.

With the enhanced memory capacity, we believe that Memformer can spark interesting works that rely on recurrence and autoregressive modeling, which will benefit tasks such as dialog and interactive systems.

